\documentclass{article}

\usepackage{arxiv}

\usepackage[utf8]{inputenc} 
\usepackage[T1]{fontenc}    
\usepackage{hyperref}       
\usepackage{url}            
\usepackage{booktabs}       
\usepackage{amsfonts}       
\usepackage{nicefrac}       
\usepackage{microtype}      
\usepackage{lipsum}		
\usepackage{graphicx}
\usepackage[numbers]{natbib}
\usepackage{doi}
\usepackage{placeins}
\usepackage{mathtools}
\usepackage{caption}
\usepackage{subcaption}
\usepackage{bm}              
\usepackage[short]{optidef}  

\title{Segmenting mechanically heterogeneous domains via unsupervised learning}

\author{ \href{https://orcid.org/0000-0002-5416-7928}{\includegraphics[scale=0.06]{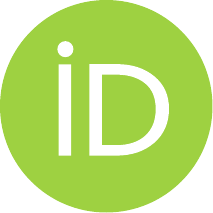}\hspace{1mm}Quan~Nguyen}\\
	Department of Mechanical Engineering\\
	Boston University\\
	Boston, MA 02215 \\
	\texttt{quan@bu.edu} \\
	\And
	\href{https://orcid.org/0000-0001-8099-3468}{\includegraphics[scale=0.06]{orcid.pdf}\hspace{1mm}Emma~Lejeune} \\
	Department of Mechanical Engineering\\
	Boston University\\
	Boston, MA 02215\\
	\texttt{elejeune@bu.edu} \\
}

\renewcommand{\shorttitle}{Segmenting mechanically heterogeneous domains via unsupervised learning}

\hypersetup{
pdftitle={A template for the arxiv style},
pdfsubject={q-bio.NC, q-bio.QM},
pdfauthor={Quan~Nguyen,Emma~Lejeune},
pdfkeywords={machine learning, soft tissue biomechanics, unsupervised learning, clustering, soft robotics},
}

\begin{document}
\maketitle

\begin{abstract}

From biological organs to soft robotics, highly deformable materials are essential components of natural and engineered systems. These highly deformable materials can have heterogeneous material properties, and can experience heterogeneous deformations with or without underlying material heterogeneity. Many recent works have established that computational modeling approaches are well suited for understanding and predicting the consequences of material heterogeneity and for interpreting observed heterogeneous strain fields. In particular, there has been significant work towards developing inverse analysis approaches that can convert observed kinematic quantities (e.g., displacement, strain) to material properties and mechanical state. Despite the success of these approaches, they are not necessarily generalizable and often rely on tight control and knowledge of boundary conditions. Here, we will build on the recent advances (and ubiquity) of machine learning approaches to explore alternative approaches to detect patterns in heterogeneous material properties and mechanical behavior. Specifically, we will explore unsupervised learning approaches to clustering and ensemble clutering to identify heterogeneous regions. Overall, we find that these approaches are effective, yet limited in their abilities. Through this initial exploration (where all data and code is published alongside this manuscript), we set the stage for future studies that more specifically adapt these methods to mechanical data.

\end{abstract}

\keywords{machine learning \and soft tissue biomechanics \and unsupervised learning \and clustering \and soft robotics}

\section{Introduction} 

\begin{figure}[h]
\centering
\includegraphics[width=0.85\textwidth]{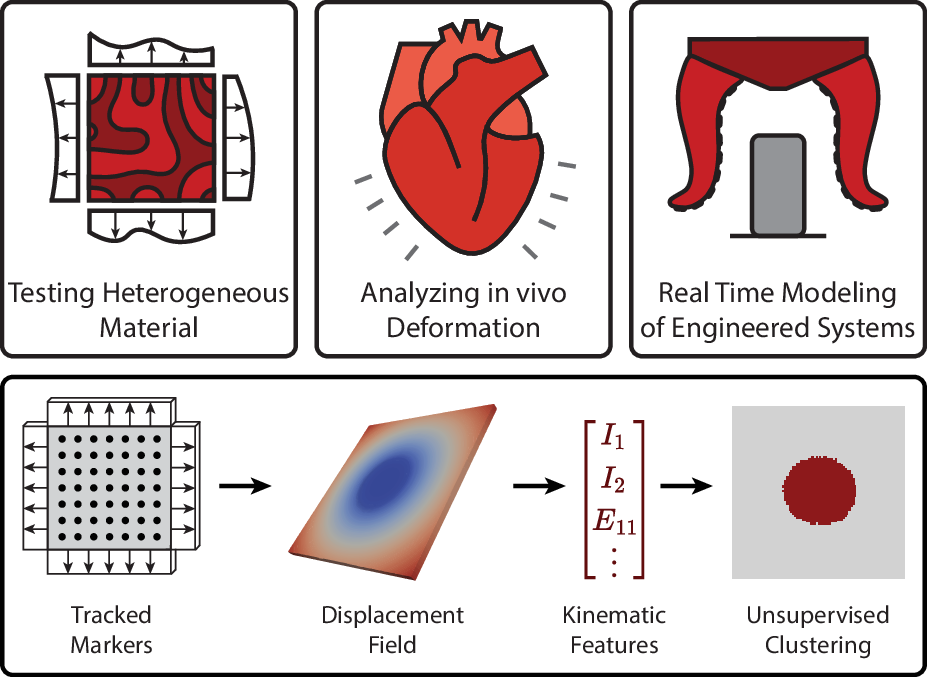}
\caption{Upper row: Motivation for our proposed study of unsupervised learning tools for analyzing soft heterogeneous materials; Lower row: Basic overview of our kinematic feature clustering pipeline.}
\label{fig:big_big_overview}
\end{figure}

From biological systems \citep{garikipati2017perspectives,rvpaper,zhang2023simulating} to engineered soft robots \citep{sensorsreview,park2012design}, highly deformable materials are ubiquitous. 
And, analyzing the mechanical behavior of common soft material natural and engineered systems poses both unique challenge and opportunity (see Fig. \ref{fig:big_big_overview}).
For example, due to the material heterogeneity inherent to many biological materials \citep{guo2022modeling,siadat2021tendon,coyle2018bio}, soft tissues often exhibit heterogeneous deformations. And, even in systems composed of homogeneous materials, we often observe heterogeneous deformations due to complex and asymmetric boundary conditions \citep{sensorsreview,svas3,wall2017method,visser2023mechanical,toaquiza2022anisotropic,de2023computational}.
Recent studies have demonstrated the suitability of computational modeling approaches for understanding and predicting the implications of material heterogeneity \citep{haugh2018investigating,sun2021platelet}, as well as interpreting observed strain fields characterized by heterogeneity \citep{linne2019data}. Notably, considerable efforts have been devoted to developing inverse analysis techniques capable of converting observed kinematic quantities (e.g., displacements, strains) into material properties and mechanical state \citep{oberai2003solution,awe,dece,chen2021learning}. However, these approaches have limitations in terms of their generalizability and reliance on precise control and knowledge of boundary conditions \citep{barbone2002quantitative}.
In this paper, we leverage the recent advances and rise in accessibility of machine learning techniques \citep{peng2021multiscale,arzani2023interpreting,chen2016combining} to analyze the kinematic behavior of soft materials undergoing large deformation.
Specifically, we consider scenarios where we can obtain the kinematics field (e.g., displacement field, strain field) of a given systems, and then perform unsupervised clustering to separate the domain into multiple sub-domains (see Fig. \ref{fig:big_big_overview}). This separation into sub-domains is useful for both analyzing the heterogeneity of biological systems \citep{witzenburg2016nonlinear}, and for allowing us to approximate the strain field in soft materials undergoing large deformation.

In the field of soft tissue mechanics, significant effort has also been made to understand the mechanically heterogeneous characteristics of soft tissues (e.g., the cornea \citep{meek2015corneal}, lumbar intervertebral discs \citep{malandrino2015poroelastic,middendorf2023lumbar}, and brain \citep{teich2021crystallinity}). Notably, the behaviors of these tissues in their healthy state is inherently complex, and certain regions of these tissues also demonstrate amplified heterogeneity when affected by diseases (e.g., breast cancer \citep{krouskop1998elastic,gemici2020relationship}, liver fibrosis \citep{sandrin2003transient}, renal cystogenesis \citep{ateshian2023computational}).
When ``full field'' data is available \citep{kennedy2017emergence,palanca2016use,you2022physics}, researchers in the field of inverse analysis have been able to recover heterogeneous material properties of these systems.
Over the past two decades, techniques have been established to perform tasks ranging from recovering the shear modulus for linear isotropic samples \citep{oberai2003solution} and non-linear samples \citep{long2022experiments} through iterative optimization , directly solving for the shear modulus \citep{awe,dece}, and solving for material properties via deep learning \citep{chen2021learning}. 
In one notable and pragmatic approach \citep{witzenburg2016nonlinear}, Witzenburg et al. introduced a method to simplify the nonlinear behavior of the whole sample and recover anisotropic material properties (i.e., preferred fiber direction, degree of anisotropy, fiber stiffness, fiber nonlinearity) by segmenting the sample into multiple partitions and assuming uniform properties within each partition.
However, despite the success of these methods and the field overall, these methods are limited by (1) high computational cost, (2) implementation challenges, and (3)  requirements and limitations on boundary conditions and related information.

Notably, the incredible functionality of biological soft tissue has helped inspire the field of soft robotics. Recently, the field of soft robotics has garnered significant attention, primarily due to the ability of soft robots to navigate difficult and extreme environments, delicately manipulate fragile objects, assist in surgical procedures, and seamlessly engage with living systems \citep{jing2022safe,calisti2017fundamentals}.
One of the grand challenges in soft robots revolves around the almost infinite degrees of freedom of soft continuum components, which results in the many distinct mechanical states \citep{sensorsreview,svas3,huang2022materials}. 
Often, a computational model is required to understand and manipulate these continuum soft robots \citep{huang2022design}. Thus, there is significant work towards reconstructing the states of these robots using sensor data \citep{wall2017method,bacher2016defsense,tapia2020makesense,spielberg2021co}. These reconstruction efforts from sensor data have raised mulitple important questions. For example: Where should we place the sensors? How many sensors are sufficient for the purpose of proprioception and/or tactile sensing?
Often, sensor placement designs with the goal of reconstructing the spatial configuration of soft robots have been developed with expert intuition \citep{wall2017method}, with expert intuition and an optimization algorithm \citep{bacher2016defsense,tapia2020makesense}, or with the help of supervised machine learning \cite{spielberg2021co}. 
Though these methods provide a reasonable reconstruction of the soft robot body, they also rely on expert intuition, an abundance of data, and/or training of a neural network. Thus, there is a strong motivation to explore alternative approaches and continue progressing this line of research for practical implementation.

In this paper, we are motivated by this broad set of applications in analyzing soft, potentially heterogeneous, and deformable materials (see Fig. \ref{fig:big_big_overview}). Here, we propose a method to separate a domain of interest into sub-domains with similar mechanical behavior. The goals of our method are: (1) to identify self-similar (i.e., homogeneous or nearly homogeneous) sub-domains within a heterogeneous domain for soft tissues applications, and (2) to identify self-similar regions within a heterogeneous strain field that can be used to reconstruct a strain field for soft robotics and related applications.
To accomplish these goals, we leverage the ability of unsupervised learning methods to uncover patterns within unlabeled data.
Typically, we evaluate the performance of unsupervised learning methods by comparing the clustering results to a ground truth.
However, due to the intricate nature of experimentally studied heterogeneous soft tissue \citep{lin2022impact,weiss2022evolving}, the ground truth for identifying material sub-domains is often absent. Therefore, in this initial methodological exploration where defining the known ground truth is essential, we will rely on computational modeling, where we can generate \emph{in silico} data with known patterns, material properties and behaviors, and boundary conditions. By creating this benchmark dataset for unsupervised learning specifically, we are able to systematically evaluate the performance of various unsupervised learning techniques, and design and implement a clustering pipeline to achieve our goals. Ultimately, we demonstrate that our clustering pipeline can both identify homogeneous sub-domains within a heterogeneous sample, and provide a baseline method for strain reconstructions in soft robotics.

The remainder of the paper is organized as follows. In Section \ref{sec:method}, we define our problem, introduce our computational dataset generation pipeline, and describe our clustering approach used to identify sub-domains. In Section \ref{sec:rnd}, we show the performance of the individual clustering methods, our ensemble clustering pipeline on heterogeneous samples, and the reconstructed strain fields using clustering. We conclude in Section \ref{sec:conclusion}. Finally, the links to the code and dataset required to reproduce our work are in Section \ref{sec:add_info}.





\section{Methods} \label{sec:method}

In this study, our goal is to better understand how unsupervised learning methods can be used (1) to identify self-similar sub-domains within a heterogeneous domain undergoing large deformations for soft tissues applications, and (2) to identify self-similar regions within a heterogeneous strain field that can be used to reconstruct a strain field for soft robotics and related applications. 
To investigate these goals, we begin in Section \ref{sec:prob_def} by defining a problem and associated assessment metrics. In Section \ref{sec:dataset}, we describe our dataset generation pipeline and subsequent open access dataset. In Section \ref{sec:kin}, we describe our procedure for extracting features from displacement fields, and in Section \ref{sec:clus_app} we describe the clustering pipeline used in this paper. The documented code to support these computational methods is provided on GitHub with example tutorials (Section \ref{sec:add_info}).

\subsection{Problem Definition and Assessment Metrics} 
\label{sec:prob_def}

To establish a dataset for investigating this problem, we define a $1\times1\times0.05$ rectangular prism domain. Then, we assign heterogeneous sub-domains from sets of heterogeneous patterns, constitutive equations, and material parameters (Section \ref{sec:dataset}, Fig. \ref{fig:overview_dataset}). The advantage of using \textit{in silico} data for this study is that we know the ground truth pattern. Given these heterogeneous domains, we apply boundary conditions and run our 3D finite element analysis (FEA) simulations to obtain a displacement field (Fig. \ref{fig:overview_dataset}). 
Then, we uniformly sample 2D displacement markers from the displacement fields, interpolate 2D regularly gridded markers, and calculate the kinematic features associated with each gridded marker (Section \ref{sec:kin}).
Our goal for this dataset is to mimic the type of full field data  (i.e., random displacement markers) that one would receive from experiments with full field imaging \citep{palanca2016use,sugerman2023speckling}.

\begin{table}[h]
\centering

\begin{tabular}{c|cccc|c}
         & $Y_1$    & $Y_2$    & $\ldots$ & $Y_s$    & $sums$   \\ \hline
$X_1$    & $n_{11}$ & $n_{12}$ & $\ldots$ & $n_{1s}$ & $a_1$    \\
$X_2$    & $n_{21}$ & $n_{22}$ & $\ldots$ & $n_{2s}$ & $a_2$    \\
$\vdots$ & $\vdots$ & $\vdots$ & $\ddots$ & $\vdots$ & $\vdots$ \\
$X_r$    & $n_{r1}$ & $n_{r2}$ & $\ldots$ & $n_{rs}$ & $a_r$    \\ \hline
$sums$   & $b_1$    & $b_2$    & $\ldots$ & $b_s$    &         
\end{tabular}
\captionsetup{skip=7pt}
\caption{Contingency table comparing the results between ground truth clusters $X=\{X_1,X_2,\ldots,X_r\}$ and clustering results $Y=\{Y_1,Y_2,\ldots,Y_s\}$ across $n$ nuber of markers. $n_{ij}$ is the number of pairs of markers in the same set for $X_i$ and $Y_j$: $n_{ij}=|X_i\cap Y_j|.$ $a$ and $b$ are the sums across the rows and columns, respectively.}
\label{table:ari}
\end{table}

\begin{figure}[h]
\centering
\includegraphics[width=1\textwidth]{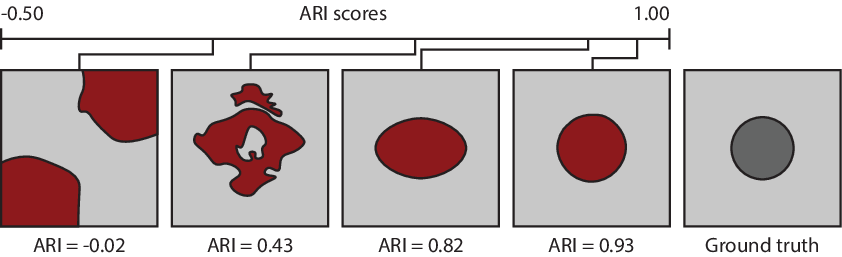}
\caption{Illustration of different clustering results for a circle inclusion domain and the respective Adjusted Rand index (ARI) score. ARI ranges from $-0.5$ to $1.0$, where a score of 0.0 indicates a result similar to random guessing, a score of 1.0 indicates the perfect clustering (i.e., matching ground truth), and a score of -0.5 indicates a result that is worse than random guessing.}
\label{fig:ARI}
\end{figure}

With this dataset defined (i.e., full field displacements and known ground truth of sub-domains locations), we can perform unsupervised clustering on the gridded markers to identify the sub-domain regions using our clustering pipeline (Section \ref{sec:clus_app}). 
Because we have known ground truth locations for heterogeneous material domain boundaries, we are able to quantitatively assess the performance of these methods via the Adjusted Rand index (ARI) \citep{ari1,ari2,scikit-learn}. Specifically, we use the ARI to compare the set of known ground truth clusters $X=\{X_1,X_2,\ldots,X_r\}$ to the set of clustering results $Y=\{Y_1,Y_2,\ldots,Y_s\}$ across $n$ number of markers.
We calculate the ARI using the contingency table (Table \ref{table:ari}) with the following the equations:
\begin{equation}
\mathrm{ARI}=\frac{\sum_{ij}\binom{n_{ij}}{2}-\bigg[\sum_i \binom{a_i}{2}\sum_j \binom{b_j}{2} \bigg] \bigg/\binom{n}{2}} {\frac{1}{2}\bigg[\sum_i\binom{a_i}{2}+\sum_j\binom{b_j}{2}\bigg]-\bigg[\sum_i\binom{a_i}{2}\sum_j\binom{b_j}{2}\bigg]\bigg/\binom{n}{2}}
\label{eqn:ARI}
\end{equation}
where $n_{ij}$ is the number of pairs of markers in the same set for $X_i$ and $Y_j$, $a_i$ and $b_j$ are obtained from Table \ref{table:ari}, and $\binom{n}{2}=n(n-1)/2$ is $n$ choose $2$. The ARI score falls in range $[-0.5,1]$, where $\mathrm{ARI}=0.0$ for random labeling, $\mathrm{ARI}=1.0$ when the clustering result is identical to the known ground truth, and $\mathrm{ARI}=-0.5$ shows that the clustering result is worse than that of random clustering (Fig. \ref{fig:ARI}).

\subsection{Computational Dataset Generation} 
\label{sec:dataset}
\begin{figure}[h]
\centering
\includegraphics[width=1\textwidth]{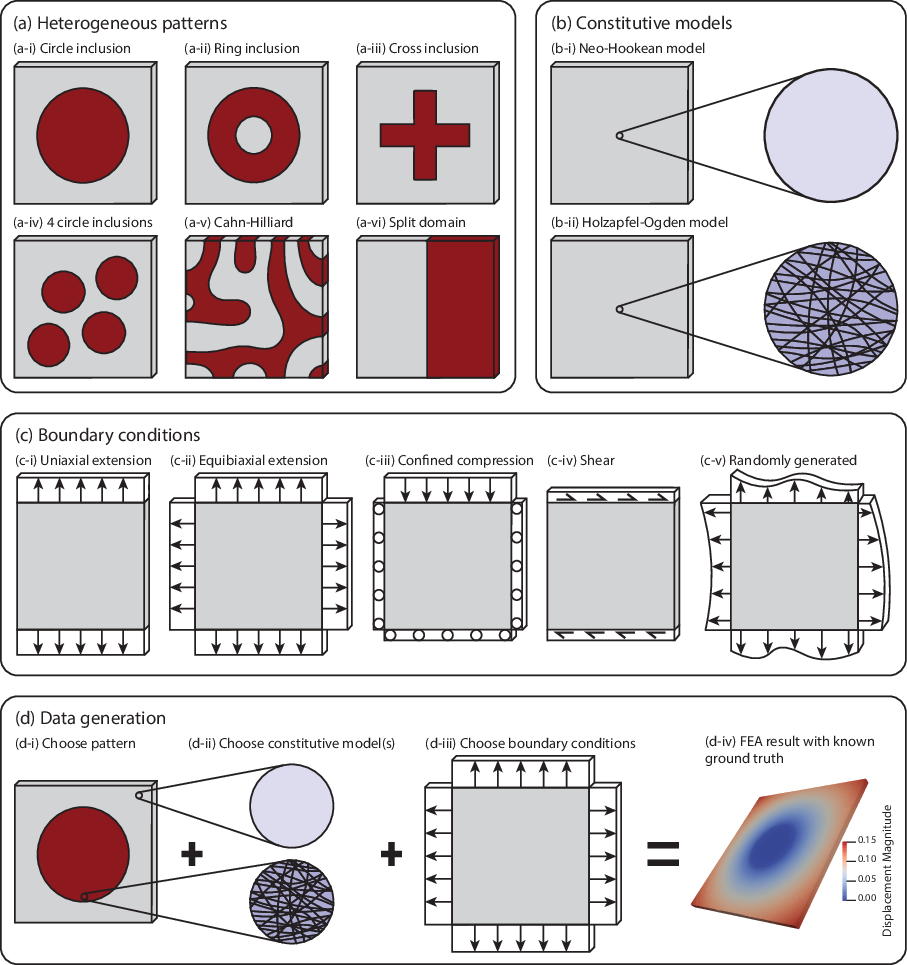}
\caption{Schematic illustration of our dataset generation process: (a) heterogeneous patterns, where the grey area and the red area depict different material domains; (b) constitutive models; (c) boundary conditions; (d) summary of our dataset generation process. Note that we can assign different constitutive models and parameters for different domains, and the example in (d-iv) is plotted in Paraview.}
\label{fig:overview_dataset}
\end{figure}

Recently, our group has published multiple benchmark datasets for evaluating machine learning methods specifically for problems in mechanics \citep{mechmnist,bic,crack,hiba,abc,ood}. However, to date, these datasets (e.g., Mechanical MNIST \citep{mechmnist}, Buckling Instability Classification \citep{bic}, Asymmetric Buckling Columns \cite{abc}) have been designed for assessing \textit{supervised} learning methods and are thus structured as large collections ($10,000+$ samples) of labeled data. Here, we introduce a new benchmark dataset designed specifically for assessing \textit{unsupervised} learning methods where the goal is to discover patterns from unlabeled data.
In this new computational dataset, referred to as Mechanical MNIST - Unsupervised Learning, we generate $6$ heterogeneous samples and report full field displacement for each sample (represented as $1000$ tracked fiducial markers \footnote{In our published dataset, we provide $\approx 1500$ tracked fiducial markers obtained from Finite Element simulations. However, for all of our analysis, we only use $1000$ tracked fiducial markers per simulation.} similar to results from digital image correlation (DIC) \citep{palanca2016use,sugerman2023speckling}). Unlike our previous datasets, which contained a \textit{small} amount of information for a \textit{large} number of samples, this dataset contains a \textit{large} amount of information for a \textit{small} number of samples thus representing an alternate class of challenges. Each sample is also accompanied by a ground truth heterogeneous material property distribution and associated metadata (e.g., constitutive models, boundary conditions) to enable: (1) assessment of method performance with respect to ground truth, and (2) extension and reproducibility of our results by others. All simulations are conducted with the Finite Element Method implemented through the Python package FEniCS \citep{fenics}. In this Section, we will elaborate on the procedure for generating this dataset, illustrated schematically in Fig. \ref{fig:overview_dataset}.

\subsubsection{Heterogeneous Patterns} 
\label{sec:heteropat}

In biological materials, there is a massive diversity in potential heterogeneous material property patterns \citep{rvpaper,witzenburg2015automatic,halvorsen2023contribution}. In this work, we focus on 6 patterns that represent distinct cases of bi-material systems. In Fig. \ref{fig:overview_dataset}a, we illustrate the six classes of patterns explored: circle inclusion, ring inclusion, cross inclusion, 4 circle inclusions, Cahn-Hilliard pattern, and split domain. 
To begin, we choose the circle inclusion used for the assessment of inverse analysis methods \ref{sec:result1}).
Then, we add the ring inclusion, where the ring is stiff, but the outer and inner backgrounds containing the ring is softer by $1$ order of magnitude for the Young's modulus.
Given that other examples lack sharp edges, we also add the cross pattern with the neo-Hookean model.
To extend the circle inclusion to simulate multiple inclusions seen in biological experiments \citep{moffitt2009formation}, we add a sample containing 4 circle inclusions with the neo-Hookean model.
Then, since the Cahn-Hilliard equation can describe some patterns in nature \citep{hiba}, we also try to identify the homogeneous sub-domains within a Cahn-Hilliard sample.
For our final pattern, we are inspired by \citep{rvpaper}, where the authors modeled a cube of right ventricular myocardial tissue with the Holzapfel-Ogden model, and divided the cube into different layers with varying fiber angle. Here, we adopt a similar strategy by designing our split pattern to be used with the Holzapfel-Ogden model, and setting the fibers angle in one half of the domain to $45^\circ$ clockwise and the other to $90^\circ$ relative to the x-axis.
In Fig. \ref{fig:overview_dataset}a, we illustrate a gray background material with a red heterogeneous pattern that will have a different set of material parameters, and potentially a different constitutive law. 
The patterns are defined as follows:

\noindent \textbf{Heterogeneous Inclusions (Circle, Ring, Cross, and 4 Circles)} 

The first bi-material system that we investigated was of heterogeneous inclusions surrounded by a homogeneous background domain. 
Within this scope, our goal was to select patterns that would pose different challenge (e.g., single vs. multiple inclusions, sharp edges) that would highlight the benefits and limitations of different domain identification methods. 
To this end, our dataset contains four distinct types of inclusion:
\begin{itemize}
    \item \textbf{Circular Inclusion}: This sample has a $1\times1\times0.05$ background with a circular inclusion in the center with a radius of $0.2$ (Fig. \ref{fig:overview_dataset}a-i).
    \item \textbf{Ring Inclusion}: This sample has a $1\times1\times0.05$ background with a ring shaped inclusion in the center, where the inner radius is $0.15$ and the outer radius is $0.35$ (Fig. \ref{fig:overview_dataset}a-ii).
    \item \textbf{Cross Inclusion}: This sample has a $1\times1\times0.05$ background with a symmetric cross inclusion in the center. The cross inclusion contains $2$ overlapping rectangles with dimensions $0.50\times0.20$, where one rectangle is rotated $90^{\circ}$ relative to the other (Fig. \ref{fig:overview_dataset}a-iii).
    \item \textbf{4-Circles Inclusion}: This sample has a background with 4-circle inclusions located at $[0.25,0.25], [0.75,0.75], [0.33,0.67], [0.67,0.33]$. The 4-circle inclusions have the same material properties with a radius of $0.125$ (Fig. \ref{fig:overview_dataset}a-iv).
\end{itemize}
From this description and from the illustrations in Fig. \ref{fig:overview_dataset}a, it should be clear that our objective was to range from \textit{simple} heterogeneous patterns where methods would likely succeed (e.g., the circular inclusion), to more complex heterogeneous patterns where methods would likely fail (e.g., delineating the inner region of the ring inclusion). \\





\noindent \textbf{Cahn-Hilliard Pattern} 

In addition to the inclusion patterns described above, we use our previously generated Cahn-Hilliard mechanical dataset \citep{hiba} as a basis for an additional pattern (Fig. \ref{fig:overview_dataset}a-v). Notably, Cahn-Hilliard patterns captures some biological phenomenon such as patterns formation \citep{liu2013phase,barrio1999two}. In soft tissue mechanics, the Cahn-Hilliard equations can qualitatively describe phenomena such as the segregation and differentiation of a tissue by its cell types (e.g., healthy to cancer cells) \citep{garikipati2017perspectives}. Here, we extrude the previously generated 2D patterns \citep{hiba} and use the two phases to delineate two heterogeneous sub-domains with different material properties. For a more detailed discussion of the Cahn-Hilliard pattern generation process, readers can refer to our previous paper \citep{hiba}.\\

\noindent \textbf{Split Domain} 

Since many biological systems are multi-layered, or have varying microstructure and mechanical properties throughout the thickness of the samples \citep{lin2022impact,sigaeva2023novel}, we want to simulate these behaviors and further test the limitations of our method in identifying these heterogeneous sub-domains. Here, we create the split domain pattern to use alongside the Holzapfel-Ogden model to simulate fibrous and layered samples with varying fiber degrees. The split domain contains two sub-domains that are adjacent to each other with size $1\times0.50\times0.05$ (Fig. \ref{fig:overview_dataset}a-vi).

\subsubsection{Constitutive Models} 
After designing the heterogeneous patterns for our synthetic data, we can further simulate the heterogeneous behaviors of soft materials by using different constitutive models at different points in space. Here, we select a common constitutive model for hyperelastic materials (i.e., neo-Hookean), and a common constitutive model for fibrous materials (i.e., Holzapfel-Ogden). 
Previously, the neo-Hookean model has been used to model hydrogels \citep{castilho2018mechanical}, and ground matrix materials \citep{mahutga2023non}.
However, since the neo-Hookean model is inadequate for describing the behavior of fibrous soft tissues \citep{guo2022modeling}, we include the Holzapfel-Ogden material model, which has been used to describe fibrous tissues such as the right ventricle of the heart \citep{rvpaper,holzapfel2009constitutive}.
In our flexible data generation pipeline, we provide the option to assign different material parameters, and different constitutive equations on the sub-domains of the same heterogeneous domain.

\noindent \textbf{Neo-Hookean}

We implement the compressible neo-Hookean constitutive model by defining the strain energy $\Psi$ as:
\begin{equation}
\Psi=\frac{\mu}{2}[\textrm{tr}(\textbf{F}^T\textbf{F})-3] - \mu \ln[\det(\textbf{F})] + \frac{\lambda}{2} \ln[\det(\textbf{F})]^2
\end{equation}
where \textbf{F} is the deformation gradient, $\lambda$ and $\mu$ are the Lam\'{e} parameters equivalent to the Young's modulus $E$ and Poisson ratio $\nu$ as $E=\mu(3\lambda+2\mu)/(\lambda+\mu)$ and $\nu=\lambda/(2(\lambda+\nu))$. For a non-fibrous heterogeneous domain, we set the Poisson's ratio $\nu=0.3$, the soft background Young's modulus $E_{soft}=1.0$, and the stiff inclusion Young's modulus $E_{stiff}=10.0$, unless specified otherwise.

\noindent \textbf{Holzapfel-Ogden}

For fibrous materials, we implement the nearly incompressible Holzapfel-Ogden constitutive model \citep{holzapfel2009constitutive}. First, we decompose the deformation gradient $\textbf{F}$ into the volumetric deformation $\textbf{F}_{vol}=J^{1/3}\textbf{I}$, and the isochoric deformation $\bar{\textbf{F}}=J^{-1/3}\textbf{F}$, where $J=\det(\textbf{F})$. The deviatoric right Cauchy-Green strain tensor is defined as $\bar{\textbf{C}}=\bar{\textbf{F}}^T\bar{\textbf{F}}$, and the isochoric invariants are as follow:
\begin{equation}
\bar{I}_1=\bar{\textbf{C}}:\textbf{I} \qquad\qquad
\bar{I}_{4f}=\textbf{f}_\textbf{0}\cdot\bar{\textbf{C}}\textbf{f}_\textbf{0} \qquad\qquad
\bar{I}_{4s}=\textbf{s}_\textbf{0}\cdot\bar{\textbf{C}}\textbf{s}_\textbf{0} \qquad\qquad
\bar{I}_{8fs}=\textbf{f}_\textbf{0}\cdot\bar{\textbf{C}}\textbf{s}_\textbf{0},
\label{eqn:iso_inv}
\end{equation}
where $\textbf{f}_\textbf{0}$ is the in-plane fiber vector, and $\textbf{s}_\textbf{0}$ is the sheet-plane vector. Then, the isochoric strain energy is defined as:
\begin{equation}
    W_{iso}=W_g+W_f+W_s+W_{fs},
\end{equation}
and the components of the isochoric strain energy are
\begin{equation}
    W_g = \frac{a}{2b}[\exp(b(\bar{I}_{1} - 3))-1],
\end{equation}
\begin{equation}
     W_f = \frac{a_f}{2b_f}[\exp(b_f(\bar{I}_{4f}-1)^2)-1],
\end{equation}
\begin{equation}
    W_s = \frac{a_s}{2b_s}[\exp(b_s(\bar{I}_{4s}-1)^2)-1],
\end{equation}
\begin{equation}
    W_{fs}= \frac{a_{fs}}{2b_{fs}}[\exp(b_{fs}\bar{I}_{8fs}^2)-1],
\end{equation}
where $a, b, a_f, b_f, a_s, b_s, a_{fs}, b_{fs}$ are material parameters. Finally, the total strain energy is defined as:
\begin{equation}
    W(C)=W_{iso}+U(J),
\end{equation}
where $U(J)=\textrm{q}[\ln(J)/J-(1/\kappa)\textrm{p}]$ applies the nearly incompressible constraint, $\kappa$ is the bulk modulus, p is the Lagrange multiplier representing the hydrostatic pressure, and q is the test function. 

\subsubsection{Boundary Conditions} 
\label{sec:bcs}
To explore the potential utility of unsupervised learning approaches in both tightly controlled laboratory conditions \citep{sree2023damage} and in less controlled scenarios (e.g., in vivo \citep{visser2023mechanical} or with complex in vitro setups \citep{blum2022tissue}), we take two approaches to implementing boundary conditions. First, we implement a suite of common boundary conditions for mechanical testing (Fig. \ref{fig:overview_dataset}c-i-iv). Then, we implement a suite of randomly generated non-standard boundary conditions that represent scenarios with limited control (Fig. \ref{fig:overview_dataset}c-v). 
For the non-standard boundary conditions, we apply the following equation to all four edges of the rectangular domain:
\begin{equation}
y=c_0+c_1x+c_2x^2+c_3x^3+c_4x^4+c_5\sin{[c_6(x-c_7)]},
\end{equation}
where $c_0, c_1, c_2, c_3, c_4, c_5, c_6, c_7$ are randomly generated constants and $x$ is the position along the edge with range $[0.0,1.0]$. All constants $c$ are selected from random uniform distributions. To add additional function shape diversity, we combine the $4^{th}$ order polynomial function with a sine function . To ensure the convergence of our FEM simulations for the randomized boundary conditions, we limit the values of $c_1, c_2, c_3, c_4$ to the range of $[-0.1,0.1]$, the values of $c_0$ to $[0.0,0.2]$, the values of $c_5$ to $[0.0,\pi/32]$, the values of $c_6$ to $[0.0,4\pi]$, and the values of $c_7$ to $[0.0,2\pi]$. The ranges for the polynomial constants further ensure that these boundary conditions lead to large deformation.

\subsubsection{Dataset Generation Pipeline}
In Fig. \ref{fig:overview_dataset}d, we schematically illustrate our dataset generation pipeline that brings together the components defined in Section \ref{sec:heteropat} - \ref{sec:bcs}.
Given these inputs, we run FEA simulations to obtain the displacement field (Fig. \ref{fig:overview_dataset}d). First, we chose a heterogeneous pattern with two domains. The two domains can have have the same constitutive model with different material parameters, or they can have different constitutive models altogether. Then, we chose the boundary condition for our simulation. Finally, we run our forward simulation in FEniCS to obtain the displacement field. The full pipeline is implemented in Python, and further details for accessing our code and the associated tutorial are available in Section \ref{sec:add_info}.

\subsection{Computing Kinematic Features from the Displacement Field} \label{sec:kin}

\begin{figure}[h]
\centering
\includegraphics[width=0.8\textwidth]{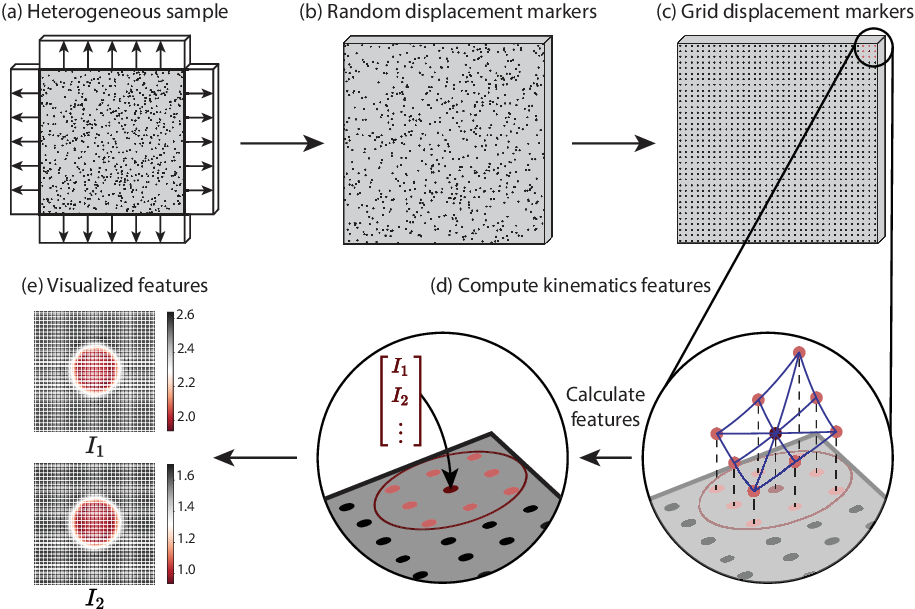}
\caption{Illustration of our feature vector calculation pipeline: (a) a heterogeneous sample with unknown materials domains; (b) randomly located marker displacements; (c) displacements converted to a grid via b-spline interpolation; (d) local neighborhood for computing local kinematic features (e.g., first and second invariants); (e) visualization of the kinematic features computed at each grid point.}
\label{fig:features}
\end{figure}

With the displacement field from our data generation pipeline, we compute the input features for our unsupervised learning clustering pipeline (Fig. \ref{fig:features}). First, we extracted randomly and uniformly sampled displacement markers from the displacement field (Fig. \ref{fig:features}b). This step replicates the results we would obtain from some experimental techniques (e.g., DIC). Then, we interpolated the uniformly sampled displacement markers via b-spline to obtain a $89\times89$ regular grid of displacement markers (Fig. \ref{fig:features}c) \citep{scikit-learn}. For instance, at each grid marker, we obtain the nearest $n$ neighbors from the random markers. Using the displacements of $n$ neighbors, we interpolated to obtain the displacement for the grid marker. We note that the randomly sampled markers and the grid markers provide similar results for our unsupervised clustering algorithms. However, the uniformly sampled markers sometimes fail to cover certain areas of the domain, whereas the grid markers allow us (1) to obtain the displacement throughout the domain using interpolation, and (2) to have displacement markers at the same positions across different boundary conditions for a given sample, which will be relevant for the implementation of ensemble clustering methods. Once we have the interpolated grid, we can compute different kinematic features based on the displacements of each marker (Fig. \ref{fig:features}d). 

\subsubsection{Kinematic features}

Given the displacement of each gridded marker, we are able to calculate mechanically relevant kinematic quantities to use as features for our clustering pipeline. For kinematic quantities with more than 1 component, we input each component as a feature (e.g., the Green-Lagrange strain tensor has 3 unique components). We can input each component as a feature for clustering). Here, we provide the list of kinematic quantities that can be used to build a feature vector, given displacement \textbf{u}:
\begin{itemize}
\item Displacement gradient: $\nabla \textbf{u}$
\item Deformation gradient: $\textbf{F}=\nabla \textbf{u}+\textbf{I}$
\item Right Cauchy-Green strain tensor: $\textbf{C}=\textbf{F}^T\textbf{F}$
\item Green-Lagrange strain tensor: $\textbf{E}=\frac{1}{2}(\textbf{F}^T\textbf{F}-\textbf{I}$)
\item First invariant: $I_1=\textrm{tr}(\textbf{C})$
\item Second invariant: $I_2=\frac{1}{2}[\textrm{tr}(\textbf{C})^2-\textrm{tr}(\textbf{C}^2)]$
\end{itemize}

\subsection{Methodological Approach to Clustering and Ensemble Clustering} \label{sec:clus_app}

In this Section, we describe multiple methods from the unsupervised learning literature. For clarity, we divide this Section into two sub-sections: basic clustering methods (i.e., \emph{k}-means clustering, spectral clustering, isolation forest, one-class support vector machine), and ensemble clustering (i.e., the cluster-based similarity partitioning algorithm). 
Here, basic clustering methods take as input a set of kinematic features, identify a latent pattern within the data, and provide a clustering result (i.e., a label for each object in the dataset). Beyond this basic approach, we are also interested in exploring scenarios where we obtain multiple sets of kinematic features, such as a sample that has a loading history that includes multiple different boundary conditions where each loading leads to a different set of kinematic features. When these richer datasets are available, we first gather multiple clustering results. Specifically, one clustering result from each set of kinematic features via basic clustering methods. Then, we use ensemble clustering to obtain a final consensus clustering result. 
To assess the performance of these unsupervised learning methods, we compare the clustering results to the known ground truth provided by our dataset (Section \ref{sec:dataset}). The results for applying the basic clustering methods are given in Section \ref{sec:result1}, and the results for the ensemble clustering method are given in Section \ref{sec:identify_het}.

\subsubsection{Basic Clustering Methods} \label{sec:basic_cluster}

\noindent \textbf{\emph{K}-means Clustering}

\emph{K}-means clustering is an extremely popular unsupervised learning method used for classification in the absence of labeled data \citep{kmeans}. Fundamentally, the \emph{k}-means clustering algorithm is designed to minimized the within-cluster sum-of-squares criterion:
\begin{equation}
    \sum^n_{i=0}\min_{\bm{\mu}_j\in \bm{C}}(\| \bm{x}_i-\bm{\mu}_j\|^2),
\end{equation}
where $n$ is the number of objects, $\bm{C}$ represents clusters, $j$ is the cluster index, $\bm{\mu}_j$ is the mean of the feature values for all objects in cluster $j$, and $\bm{x}_i$ represents feature values for the $i^{th}$ object. Here, we used scikit-learn \citep{scikit-learn} to implement \emph{k}-means clustering and cluster our input sets of kinematics features into a corresponding sets of labels.

\noindent \textbf{Spectral Clustering}

Spectral clustering is an unsupervised learning method based on calculating the normalized Laplacian of features describing the objects, and clustering the largest eigenvectors of the normalized Laplacian to obtain the labels for all objects \citep{shi2000normalized}. Specifically, given a set of data points $\textbf{x}=\{\bm{x}_1,\bm{x}_2,\ldots,\bm{x}_n\}$ with $n$ as the number of data points, we compute a $n\times n$ affinity matrix \textbf{A} \footnote{The terms ``affinity matrix'' and ``similarity matrix'' are used interchangeably in this literature.} describing the relations between the data points using the Gaussian radial basis function with the Euclidean norm:
\begin{equation}
    \textbf{A}_{ij}=k(\textbf{x}_i, \, \textbf{x}_j) \, = \, \textrm{exp}(-\gamma \, ||\textbf{x}_i-\textbf{x}_j||^2) \, ,
\end{equation}
where $\gamma=1.0$. Alternatively, we can also input a pre-computed affinity matrix to spectral clustering, and continue the remaining steps. Then, we define the random walk normalized graph Laplacian matrix \citep{shi2000normalized,von2007tutorial} as:
\begin{equation}
\textbf{L}_{rw}\coloneqq \textbf{I}-\textbf{D}^{-1}\textbf{A},
\end{equation}
where \textbf{I} is the identity matrix, and \textbf{D} is the diagonal matrix obtained via $\textbf{D}_{ii}=\sum^n_{j=1}\textbf{A}_{ij}$. Then, we compute the $k$ largest eigenvectors for $\textbf{L}_{rw}$, where $k$ is the hyperparameter representing the number of clusters. To obtain the clustering labels, we cluster via \emph{k}-means the projections of the data points onto the $k$ largest eigenvectors.
Here, we used spectral clustering through scikit-learn \citep{scikit-learn} to (1) identify self-similar sub-domains within our heterogeneous domains undergoing a single boundary condition, and (2) partition a similarity matrix as part of our ensemble clustering pipeline (details in Section \ref{sec:ens_cluster}).


\noindent \textbf{Isolation Forest}

Isolation forest (iForest) is an anomaly detection method, which classifies input data into a ``normal'' group and an ``abnormal'' group \citep{liu2008isolation}. 
Given a set of data points alongside their features, the iForest algorithm works by recursively selecting a random feature and an arbitrary threshold, then splitting the data based on the randomly selected feature and threshold. Following this splitting process, the data points that require less splitting  before being isolated from the rest of the data (i.e., an isolated data point is the only data point in a group) are labeled ``abnormal.'' Here, we use the scikit-learn package \citep{scikit-learn} to identify the different regions within our heterogeneous domains by considering the ``normal'' and ``abnormal'' data points as two different clusters.

\noindent \textbf{One-class Support Vector Machine}

The one-class Support Vector Machine is another anomaly detection method based upon the common supervised learning method Support Vector Machines (SVM). Briefly, given a training dataset $(\bm{x}_1, \, y_1),\ldots,(\bm{x}_n, \, y_n)$, where $x$ is the input and $y$ the binary output with values $-1$ or $1$, the linear hard-margin SVM approach finds the maximum-margin hyperplane dividing the two classes \citep{cortes1995support,vapnik2006estimation}. Assuming that the training data is linearly separable, we are able to find 2 parallel margins that separate the 2 classes of data, and the maximum-margin hyperplane is the hyperplane lying halfway between the 2 margins. Here, we obtain the 2 margins for a set of points through the optimization problem:
\begin{mini!}
    {\textbf{w},b}{||\textbf{w}||^2_2} {\label{eqn:svm}}{}
    \addConstraint{y_i(\textbf{w}^T\textbf{x}_i-b)}{\geq 1 \qquad \forall i\in \{1,\ldots,n\}}
\end{mini!}
where \textbf{w} is the normal vector to the hyperplane, $b$ the bias term, and $n$ the total number of data points. 
With the one-class SVM, the data will be unlabeled. Thus, instead of finding a maximum-margin hyperplane, one-class SVM finds a sphere with minimum volume describing a region in the feature space that contains the majority of the unlabeled data points. Then, the points outside of this spherical region are considered ``abnormal'' data points \citep{tax1999support}. However, since most data are not spherically distributed, we map the input data to a feature space more suitable for a spherical boundary via a Gaussian radial basis function with the Euclidean norm:
\begin{equation}
    k(\textbf{x}_i,\textbf{x}_j)=\textrm{exp}(-\gamma ||\textbf{x}_i-\textbf{x}_j||^2),
\end{equation}
where $\gamma=1/(d\,\textrm{Var}(\textbf{x}))$, $d$ is the dimension of \textbf{x}, and Var(\textbf{x}) is the variance of \textbf{x}. The resulting $n\times n$ feature space \footnote{This feature space is also referred to as the ``Gram matrix'' in the literature.} represents the similarity between a data point and all other data points through the Gaussian radial basis function.
Here, we use one-class SVM through the scikit-learn package \citep{scikit-learn} to identify the different regions within our heterogeneous domains.

\subsubsection{Clustering Pipeline using an Ensemble Clustering Method} \label{sec:ens_cluster}

\begin{figure}[h]
\centering
\includegraphics[width=1\textwidth]{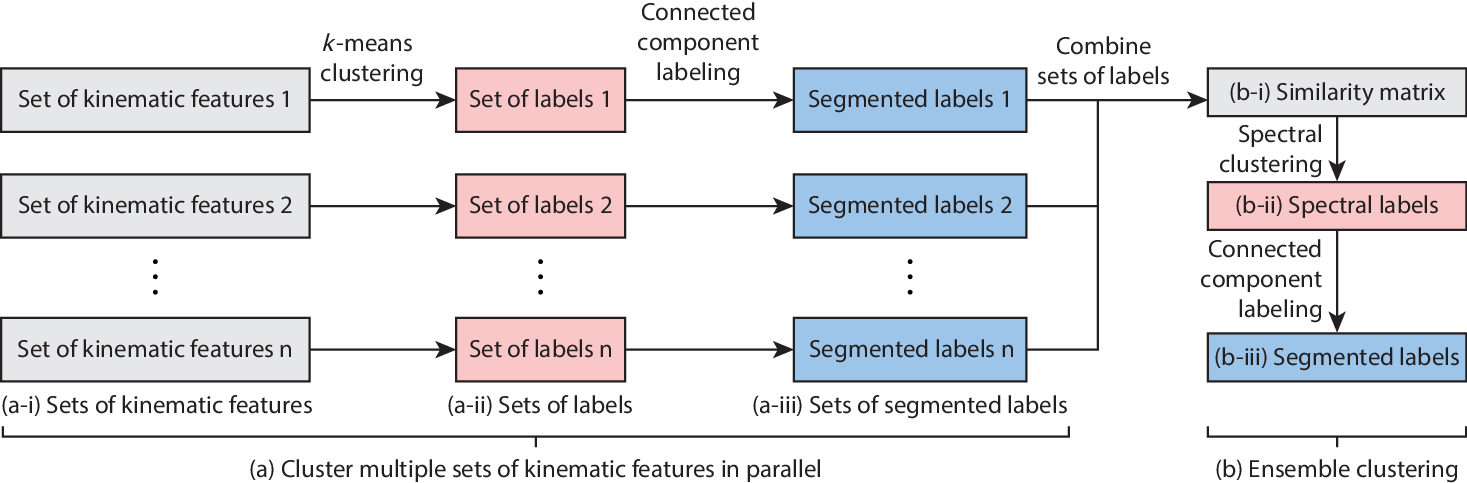}
\caption{Conceptual illustration of our ensemble clustering pipeline: (a-i) sets of kinematic features obtained from Fig. \ref{fig:features}; (a-ii) sets of labels obtained from \emph{k}-means clustering; (a-iii) sets of positional segmented labels obtained from connected component labeling; (b-i) similarity matrix obtained by combining all sets of labels; (b-ii) set of labels obtained from spectral clustering; (b-iii) final segmented labels obtained from connected component labeling.}
\label{fig:pipeline}
\end{figure}

When there are multiple sets of kinematic features for each domain (e.g., multiple different applied boundary conditions for the same sample, see Section \ref{sec:kin}) we are able to further identify self-similar sub-domains using ensemble clustering.
Here, we propose a clustering pipeline, which includes the use of basic clustering methods (\emph{k}-means clustering, spectral clustering),  an ensemble clustering method (the cluster-based similarity partitioning algorithm), and filtering techniques from image analysis (connected component labeling, cluster size thresholding). This pipeline is schematically illustrated in Fig. \ref{fig:pipeline}. 

To implement this pipeline, we first cluster the sets of kinematic features - describing the regularly gridded markers spanning the domain of the samples - through \emph{k}-means clustering to obtain multiple sets of labels. There will be one set of labels per applied boundary condition. Then, since the sets of labels correspond to the regularly gridded markers, we convert each set of labels into an image-like array, where each pixel represents a grid marker with the pixel value as the label. With the labels converted to images, we segment these images using a connected component labeling algorithm to separate the different clusters by position. 
At the end of this segmentation step, we might have some very small clusters (i.e., clusters with fewer than $5$ data points), so we impose a minimum cluster size thresholding and re-assign these small clusters to the nearest large cluster. 
After performing \emph{k}-means clustering and filtering on the sets of kinematics features, we have multiple sets of segmented labels (Fig. \ref{fig:pipeline}a), which is suitable for ensemble clustering methods.

Here, we perform ensemble clustering via a Cluster-based Similarity Partitioning Algorithm (CSPA). Specifically, we combine the sets of segmented labels into a similarity matrix, and partition the similarity matrix into clusters using spectral clustering (the process for constructing the similarity matrix and CSPA is defined below). Then, we again segment the ensemble results via connected component labeling and cluster size thresholding to obtain the final clustering result (Fig. \ref{fig:pipeline}b). Overall, our clustering pipeline identifies the self-similar sub-domains by clustering the sets of kinematic features, then segment the sub-domains again by position to obtain the final result. In this manner, the sub-domains are self-similar in terms of both their mechanical behaviors, and their positions in space.

\begin{table}[h]
\centering
\begin{tabular}{l|llll}
      &                 &                 &                 &                \\
      & $\lambda^{(1)}$ & $\lambda^{(2)}$ & $\lambda^{(3)}$ & $\lambda^{(4)}$ \\ \hline
$x_1$ & $1$             & $2$             & $1$             & $1$             \\
$x_2$ & $1$             & $2$             & $1$             & $2$             \\
$x_3$ & $1$             & $2$             & $2$             & ?               \\
$x_4$ & $2$             & $3$             & $2$             & $1$             \\
$x_5$ & $2$             & $3$             & $3$             & $2$             \\
$x_6$ & $3$             & $1$             & $3$             & ?               \\
$x_7$ & $3$             & $1$             & $3$             & ?              
\end{tabular}
$\ \ \Leftrightarrow\ \ $
\begin{tabular}{l|lll|lll|lll|ll}
 & $H^{(1)}$ &       &       & $H^{(2)}$ &       &       & $H^{(3)}$ &       &       & $H^{(4)}$ &        \\
 & $h_1$     & $h_2$ & $h_3$ & $h_4$     & $h_5$ & $h_6$ & $h_7$     & $h_8$ & $h_9$ & $h_{10}$    & $h_{11}$ \\ \hline
$x_1$ & $1$       & $0$   & $0$   & $0$       & $1$   & $0$   & $1$       & $0$   & $0$   & $1$       & $0$    \\
$x_2$ & $1$       & $0$   & $0$   & $0$       & $1$   & $0$   & $1$       & $0$   & $0$   & $0$       & $1$    \\
$x_3$ & $1$       & $0$   & $0$   & $0$       & $1$   & $0$   & $0$       & $1$   & $0$   & $0$       & $0$    \\
$x_4$ & $0$       & $1$   & $0$   & $0$       & $0$   & $1$   & $0$       & $1$   & $0$   & $1$       & $0$    \\
$x_5$ & $0$       & $1$   & $0$   & $0$       & $0$   & $1$   & $0$       & $0$   & $1$   & $0$       & $1$    \\
$x_6$ & $0$       & $0$   & $1$   & $1$       & $0$   & $0$   & $0$       & $0$   & $1$   & $0$       & $0$    \\
$x_7$ & $0$       & $0$   & $1$   & $1$       & $0$   & $0$   & $0$       & $0$   & $1$   & $0$       & $0$   
\end{tabular}

\captionsetup{skip=7pt}
\caption{Hypergraph construction with different sets of labels $\lambda$. The left table contains the original $4$ sets of labels $\lambda$ and $7$ objects $x$. The numbers denote the clusters assigned to objects, and ? indicates that the objects has no assigned cluster. The right table contains the hypergraph with $11$ hyperedges, $7$ objects, and the binary value indicates whether an object belong to the hyperedge. Adapted from \citep{cluster_en}.}

\label{table:hypergraph}

\end{table}

\noindent \textbf{Cluster-based Similarity Partitioning Algorithm}

The Cluster-based Similarity Partitioning Algorithm is an ensemble clustering method, where the information from multiple sets of labels is aggregated to obtain a final consensus set of labels \citep{cluster_en}.
Specifically, given multiple sets of clustering labels $\lambda$, CSPA first converts these labels into a hypergraph $H$ (Table \ref{table:hypergraph}). 
Here, a hypergraph $H$ is a binary matrix representation of multiple sets of clustering labels, where each row represents an object to be clustered, each column represents a group for each set of labels, and the binary value indicates whether an object belongs to a group. 
For example, the set of labels $\lambda^{(1)}$ has 3 groups (left side of Table \ref{table:hypergraph}), which corresponds to 3 columns (i.e., $h_1,h_2,h_3$) in hypergraph $H$ with binary values assigning a group to the objects (right side of Table \ref{table:hypergraph}). 
After setting up the hypergraph $H$, we compute the similarity matrix $S=\frac{1}{r}HH^T$, where $r$ is the number of sets of clustering labels.
Then, we partition the similarity matrix by using spectral clustering as a graph-partitioning algorithm \citep{shi2000normalized,von2007tutorial}. Specifically, we input the similarity matrix as an affinity matrix into spectral clustering using the scikit-learn package \citep{scikit-learn}, and obtain the final consensus labels.

\section{Results and Discussion} \label{sec:rnd}

In this Section, we examine the performance of our clustering method described in Section \ref{sec:clus_app} on the dataset described in Section \ref{sec:dataset}. We begin in Section \ref{sec:result1} by examining a wide range of clustering methods and kinematic features on the "circle inclusion" example and use the results of this study to down-select feature options.  Then, in Section \ref{sec:identify_het}, we discuss the results of our clustering pipeline for identifying homogeneous sub-domains within a heterogeneous domain, on both controlled boundary conditions and poorly controlled random boundary conditions. Finally, in Section \ref{sec:sensors}, we discuss using our clustering pipeline to obtain reconstructed representations of strain fields for applications in sensor placement.

\subsection{\emph{K}-means, spectral clustering, and invariants of the right Cauchy-Green strain tensor provide the best clustering performance for individual boundary conditions.} \label{sec:result1}

As introduced in Section \ref{sec:clus_app}, there are multiple options for both clustering algorithms (i.e., \emph{k}-means clustering, spectral clustering, iForest, one-class SVM) and kinematic features (i.e, displacement $\textbf{u}$, deformation gradient $\textbf{F}$, first and second invariants $I_1\textrm{ and }I_2$). To narrow down these options, we assessed the performance of the $3$ kinematic features and $4$ clustering methods on a neo-Hookean sample with a circle inclusion pattern. The circular inclusion pattern, where a circle with radius $0.2$ is centered in a square domain with side length $1.0$, is based on a common test problem from the inverse analysis literature \citep{awe,dece,ray2022efficacy}. To generate this initial dataset, we follow the process in Figure \ref{fig:overview_dataset}d, and select the circle inclusion pattern, neo-Hookean constitutive model, and $3$ different boundary conditions. For each boundary condition, we obtain the displacements $\textbf{u}$, and calculate the deformation gradient $\textbf{F}$ and invariants of the right Cauchy-Green strain tensor $\textbf{C}$ (process from Figure \ref{fig:features}). 
Then, we set up the input kinematic feature vectors with physical meaning (i.e., displacements, deformation gradient, and invariants). 
Given these kinematic feature vectors, we then cluster them to identify mechanical sub-domains. In this initial study, we explored two popular methods from the unsupervised learning literature - \emph{k}-means clustering and spectral clustering, and two popular methods from the unsupervised anomaly detection literature - iForest and one-class SVM, all defined in Section \ref{sec:clus_app}. Finally, we evaluate our result using the ARI score defined in Section \ref{sec:prob_def}. 

\begin{figure}[h]
\centering
\includegraphics[width=1\textwidth]{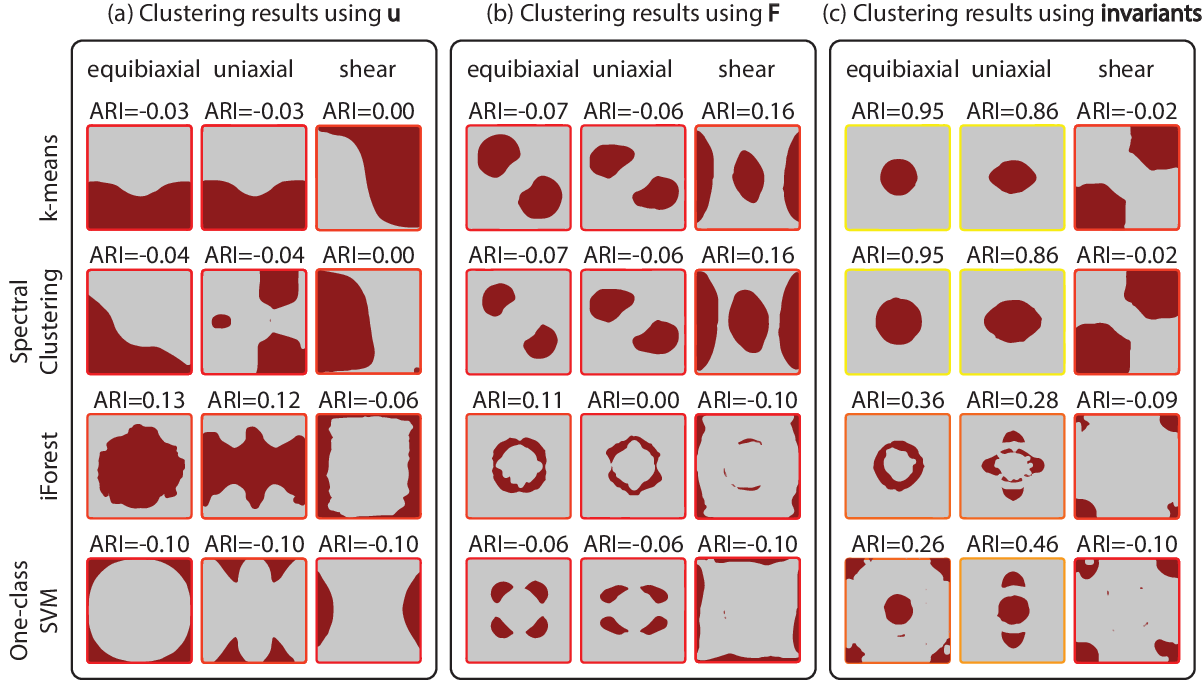}
\caption{Clustering results for our circle inclusion pattern using different kinematic features: (a) clustering results using the displacement $\textbf{u}$; (b) clustering results using the deformation gradient $\textbf{F}$; (c) clustering results using invariants of the right Cauchy-Green $\textbf{C}$. The best clustering results are shown in yellow boxes, while the poorer clustering results are in orange and red boxes.}
\label{fig:invariants}
\end{figure}

In Fig. \ref{fig:invariants}, we show the results of this initial investigation. 
Out of the $3$ options for kinematics features, the invariants of the right Cauchy-Green strain tensor consistently provide the highest ARI score when evaluated using our known ground truth material property distribution (Fig. \ref{fig:invariants}c). 
Though the unsupervised clustering methods that we investigated here are all data type agnostic, and thus do not account for the physical interpretations of our data, this result is consistent with physical intuition.
Specifically, since the first and second invariants represent the sum of the stretch and the determinant of the right Cauchy-Green strain tensor, respectively, we expect the sub-domains with different material properties to have significantly different values for the invariants. In other words, the sub-domains with higher stiffness should have a lower stretch and change in deformation, so the invariants of the sub-domains should have lower values compared to that of the sub-domains with lower stiffness. 

Out of the $4$ clustering methods investigated, \emph{k}-means and spectral clustering led to the highest ARI scores, which means that these clustering methods were able to best identify the circular inclusion sub-domain. Since the boundary conditions in the experimental settings might not be perfectly equibiaxial or uniaxial, we also provide the performance of \emph{k}-means for a circle inclusion with boundary conditions varying from equibiaxial to biaxial to uniaxial in Appendix \ref{app:varying_bcs}. The high performance of \emph{k}-means and spectral clustering for clustering mechanical data has also been recently observed when identifying grain boundaries in poly-crystalline materials \citep{linne2019data}, and identifying damage mechanisms in acoustic emission \citep{muir2021machine,muir2023quantitative}. 
And, similar to approaches from the inverse analysis literature \citep{dece,awe_elasticity}, we were able to recover the circle inclusion. However, unlike our method, the inverse analysis methods proposed in \citep{dece,awe_elasticity} also predict the shear modulus $\mu$, given the strain field and either the shear modulus at a point or the mean shear modulus over the domain. 
In comparing unsupervised learning to inverse analysis, it is important to note the clear trade off between ease and accessibility of implementation and fidelity of outcomes. On one hand, unsupervised learning approaches only requires either the displacement/strain field or displacement markers over the domain and are easily implemented through the Scikit-learn python package \citep{scikit-learn}. On the other hand, unsupervised learning approaches cannot directly recover the shear modulus alongside the location of the sub-domains. Thus, we anticipate that unsupervised learning approaches will be most relevant when the main goal is to identify self-similar sub-domains within a heterogeneous domain.

\subsection{Clustering and ensemble clustering can be used to identify heterogeneous material property distributions within a domain.} \label{sec:identify_het}

One of the goals of implementing the ensemble clustering pipeline is to identify homogeneous sub-domains located within a heterogeneous domain. For instance, in Section \ref{sec:result1}, we previously identified a stiff circle inclusion from a soft background, where the sample deformed via equibiaxial extension. Here, we ensure the robustness of our pipeline by accessing its performance on different patterns, constitutive models, and boundary conditions.
First, we create synthetic data with a known ground truth material property distribution via our data generation process (Fig. \ref{fig:overview_dataset}d). Then, for each example, we cluster the heterogeneous domains with our clustering pipeline (Fig. \ref{fig:pipeline}). For this analysis, we use $6$ heterogeneous patterns (circle inclusion, ring inclusion, cross inclusion, 4 circle inclusions, Cahn-Hilliard pattern, and split domain), $2$ constitutive models (neo-Hookean, and Holzapfel-Ogden), and $4$ boundary conditions (equibiaxial extension, uniaxial extension, shear, and confined compression). 

\begin{figure}[!t] 
\includegraphics[width=1\textwidth]{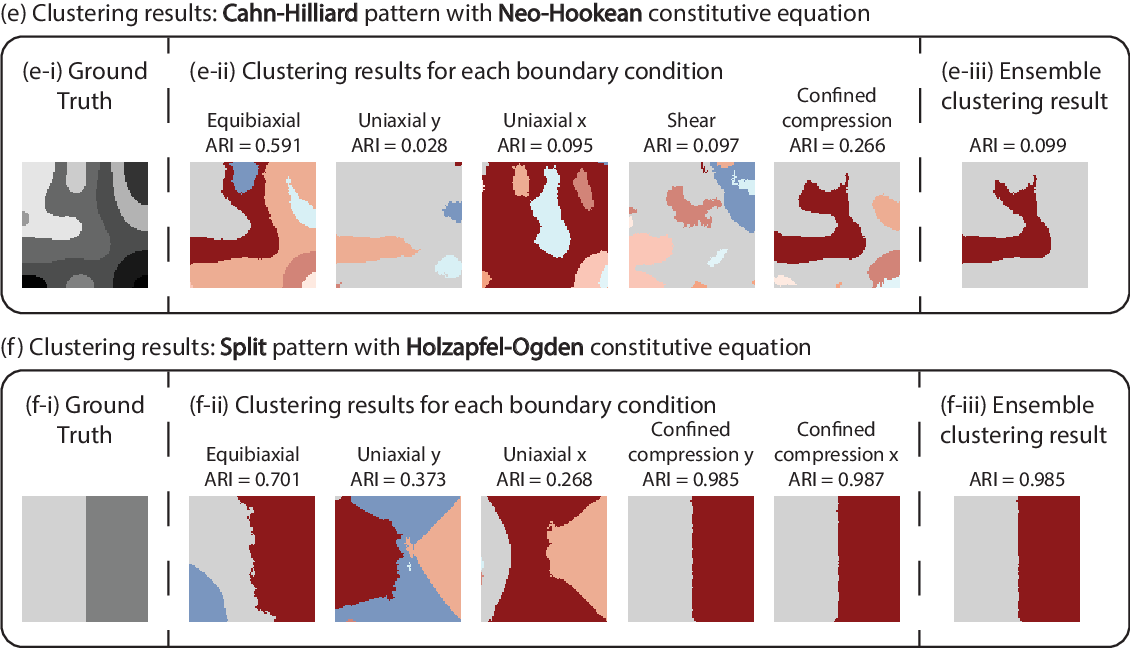}
\caption{Clustering results for different patterns and constitutive equations combinations: (a) circle inclusion pattern with Neo-Hookean constitutive equation; (b) ring inclusion pattern with Neo-Hookean constitutive equation; (c) cross inclusion pattern with Neo-Hookean constitutive equation; (d) 4 circle inclusions pattern with Neo-Hookean constitutive equation;  (e) Cahn-Hilliard pattern with Neo-Hookean constitutive equation; (f) split pattern with Holzapfel-Ogden constitutive equation. For all 6 cases, (i) is the ground truth, (ii) shows the clustering results for sets of kinematic features generated from different boundary conditions, and (iii) shows the ensemble clustering result.}
\label{fig:singlevsensemble}
\end{figure}

After data generation, we obtain the kinematic features at each grid marker (following the process illustrated in Fig. \ref{fig:features}), and cluster the markers into homogeneous sub-domains. Following the results from Section \ref{sec:result1}, we use \emph{k}-means and spectral clustering with the number of clusters $k$ set to $2$, and invariants of the right Cauchy-Green strain tensors as the chosen kinematic features to identify the different sub-domains. 
To show the robustness of our ensemble clustering pipeline, we compare the sub-domains identified in $2$ cases: the sub-domains identified when considering only $1$ boundary condition at a time, and the sub-domains identified when considering multiple boundary conditions at once (Fig. \ref{fig:singlevsensemble}).
When we consider multiple boundary conditions together, we are performing ensemble clustering for all the neo-Hookean samples with the exception of the Cahn-Hilliard example (Fig. \ref{fig:singlevsensemble}a-e), the ensemble clustering result provides a similar or better result than the individual clustering results for each boundary condition. While the individual boundary condition clustering results fail to identify the inclusions in many cases, the ensemble results recovers the different sub-domains with a high accuracy (i.e., high ARI score when evaluated against the ground truth). However, for the samples with more than $2$ disconnected sub-domains (i.e., ring inclusion, 4 circle inclusions, Cahn-Hilliard), we can only identify the homogeneous sub-domains, but we fail to determine whether one sub-domain has the same material properties as another sub-domain. Overall, our ensemble clustering pipeline performs worst in the Cahn-Hilliard sample, and the equibiaxial case outperforms all other clustering results by a large margin. We suspect that for samples with many similarly sized homogeneous sub-domains (i.e., not inclusions), our ensemble pipeline may function poorly.
For our split domain Holzapfel-Ogden case, we observe that the results for the confined compression cases provide the best result, alongside the ensemble clustering. However, in biological settings, the fiber distribution typically varies across the domain \citep{rvpaper}, instead of distinctively split into $2$ sub-domains, which means that our method might have issues identifying the sub-domains when encountering fibrous samples in real applications.

\begin{figure}[!b]
\centering
\includegraphics[width=1\textwidth]{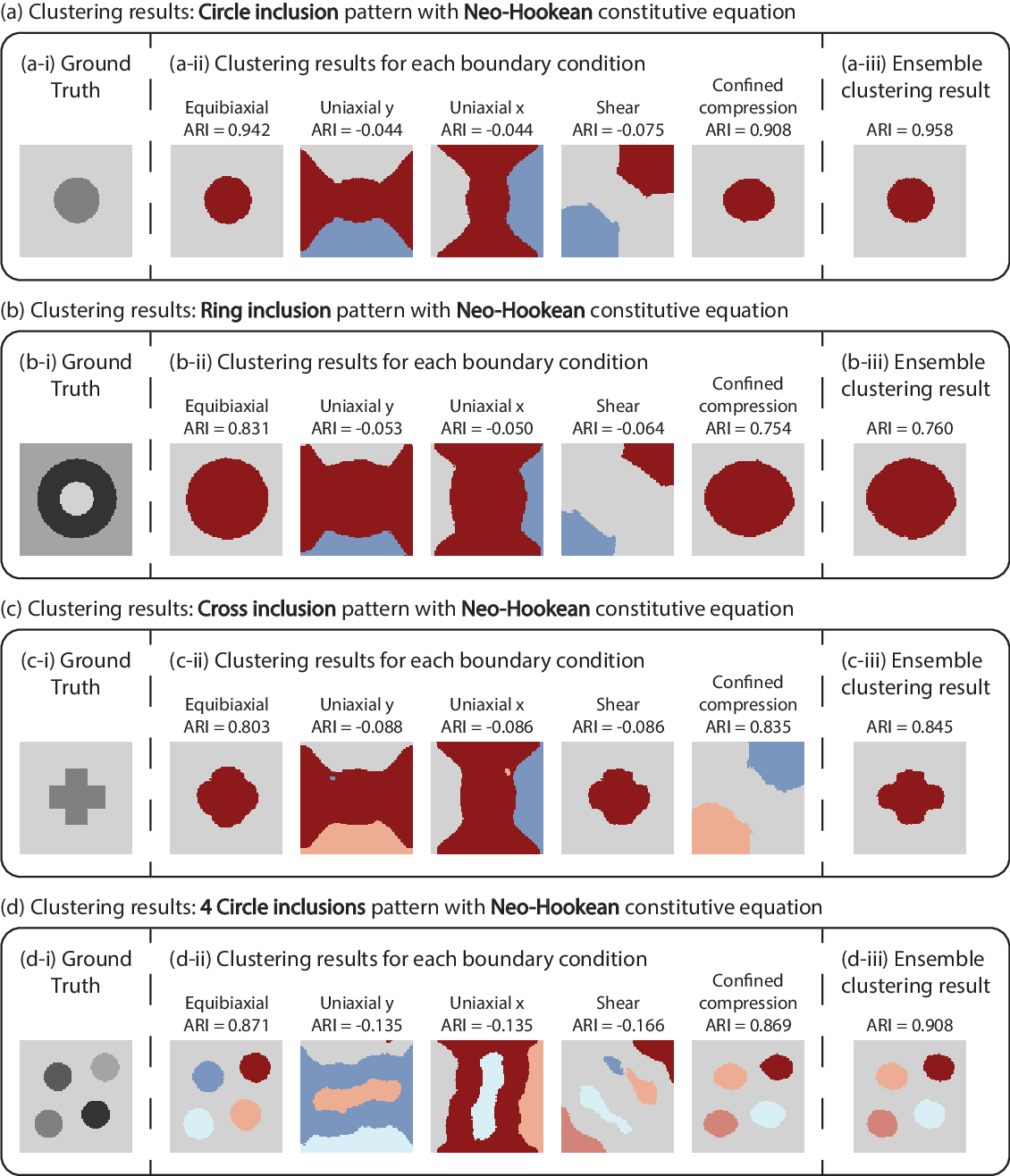}
\end{figure}

\begin{figure}[p]
\centering
\includegraphics[width=0.9\textwidth]{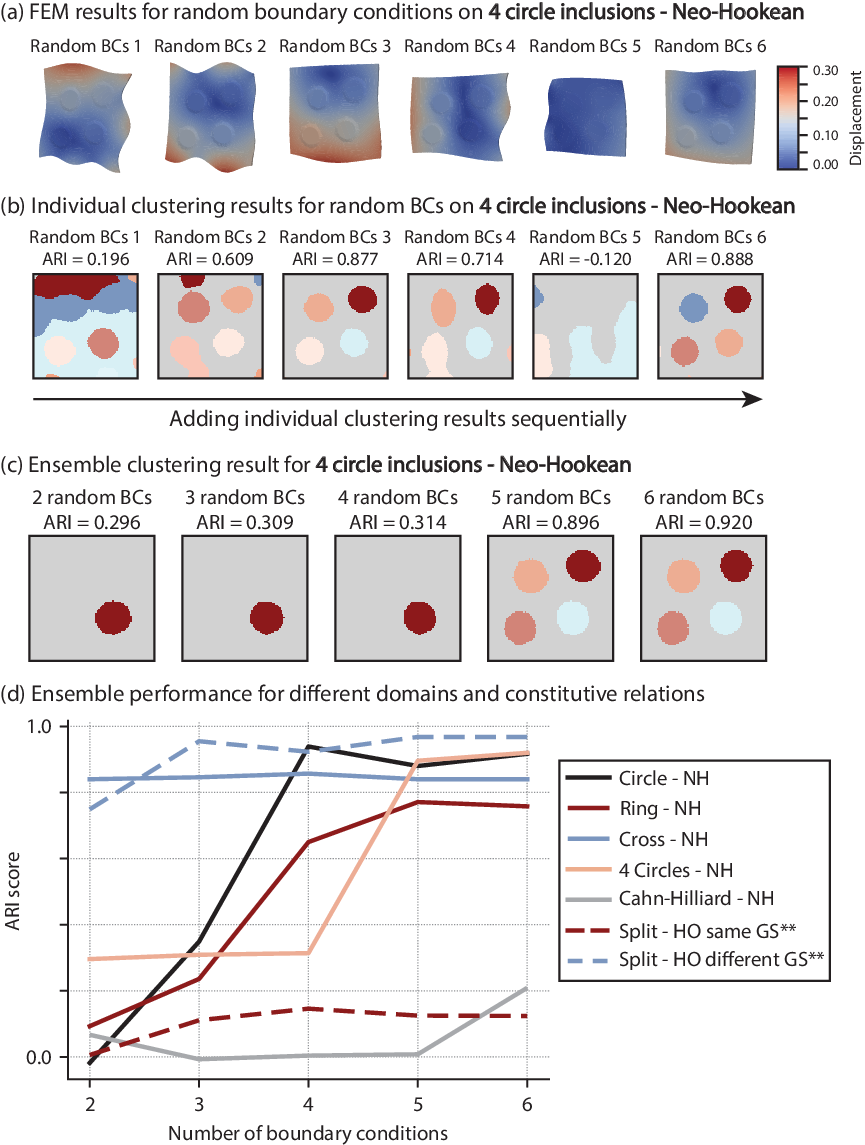}
\caption{Ensemble clustering results for random boundary conditions: (a) FEM simulation results for random boundary conditions on a 4 circle inclusions - Neo-Hookean constitutive relation sample; (b) individual clustering results for random boundary conditions on a 4 circle inclusions - Neo-Hookean sample; (c) ensemble clustering result for random boundary conditions on the 4 circle inclusions - Neo-Hookean sample as the number of boundary conditions increases; (d) ensemble clustering results for random boundary conditions for various domain and constitutive relation samples. ** Here, GS refers to "ground substance".}
\label{fig:random_bcs}
\end{figure}

While our clustering pipeline works well for identifying inclusions within a domain undergoing common experimental boundary conditions, it is not necessarily possible to obtain such tightly controlled applied loads in all scenarios. 
For example, during in vivo loading, a tissue domain may undergo multiple different load conditions that are poorly understood. Here, we simulate $7$ samples with different combinations of heterogeneous patterns and constitutive models undergoing \emph{random} boundary conditions (defined in Section \ref{sec:bcs}). 
The first $6$ samples are similar to the ones above: $5$ samples using the neo-Hookean constitutive models (i.e., circle inclusion, ring inclusion, cross inclusion, 4 circle inclusions, and Cahn-Hilliard pattern), and $1$ sample using the Holzapfel-Ogden constitutive model on the split domain with the sub-domains having the same ground substance stiffness and different fiber angles (i.e., $45^\circ$ clockwise and $90^\circ$ clockwise relative to the x-axis). 
Additionally, we include $1$ more sample using the Holzapfel-Ogden alongside the split domain, but both the ground substance stiffness and fiber angles for the sub-domains are different (i.e., one sub-domain has double the stiffness value compared to the other). 
In the previous example with the common boundary conditions, we were able to identify the sub-domains with high accuracy. However, we needed information from $5$ different boundary conditions, which might not be available in many settings. Here, we perform ensemble clustering on random boundary conditions, and assess the performance of the ensemble as the number of random boundary conditions increases. Broadly speaking, we anticipate that as more information is available to our ensemble, our ensemble clustering result should improve. 

In Fig. \ref{fig:random_bcs}a, we simulate the 4 circle inclusions - neo-Hookean sample with $6$ different \emph{random} boundary conditions \footnote{Random boundary conditions are provided with seed numbers for reproducibility.}, and we perform clustering on the kinematic features generated from the individual random boundary conditions (Fig. \ref{fig:random_bcs}b). Since we want to understand the performance of our ensemble method as we obtain more data from different boundary conditions, we first perform ensemble clustering with only $2$ boundary conditions. Then, we increase the number of boundary conditions until we obtain a good clustering result for the 4 circle inclusions - neo-Hookean sample (Fig. \ref{fig:random_bcs}c). We found that after $5$ random boundary conditions, our ensemble pipeline successfully identifies the 4 circle inclusions.
Aside from this 4 circle inclusion sample, we also test our ensemble pipeline on a variety of samples. For our neo-Hookean samples, as expected, we observe a positive trend between the number of boundary conditions in our ensemble and the ARI score (Fig. \ref{fig:random_bcs}d), which means that our ensemble clustering pipeline still performs well in this scenario.
However, the clustering result for the Cahn-Hilliard sample remains poor. Based on this result, we expect our clustering pipeline to provide less accurate results for domains with more complicated patterns, particularly when patterns are not ``inclusions''.
For the Holzapfel-Ogden samples where the sub-domains have the same ground substance stiffness, our ensemble clustering pipeline fails to identify the sub-domains for both the circle inclusion and the split domain patterns (i.e., ARI score $< 0.30$). 
However, when we consider the case where the sub-domains of the fibrous samples have different ground substance stiffness, we find that our ensemble pipeline can identify the sub-domains where the ground substance stiffness is different with an ARI score $\geq 0.70$.

Overall, our ensemble clustering pipeline successfully identifies the sub-domains for a variety of heterogeneous patterns, and under different boundary conditions. Our clustering pipeline works well for heterogeneous inclusions where the sub-domains having different stiffness, while the pipeline performs worse for more complicated patterns (i.e., Cahn-Hilliard), and for sub-domains with no difference in their stiffness. These results show that when information from loading with multiple different boundary conditions is available, ensemble clustering is a viable method for identifying inclusions in heterogeneous domains.

\subsection{Clustering and ensemble clustering can be used to create reconstructed strain fields.} \label{sec:sensors}

In addition to identifying heterogenous material properties, clustering can be used to identify self-similar regions within a domain \citep{liu2016self,Ferreira2023}. Abstractly, this is consistent with work from the computer vision literature on clustering as a form of data compression \citep{paek2015k}. And, in the context of speeding up finite element analysis simulations, this practice is one approach to approximating heterogeneous material behavior \citep{liu2016self,Ferreira2023}. Practically, in physical systems, we can also use clustering approaches to prescribe sensor positions when there are a limited number of sensors available to approximate a heterogeneous strain field \citep{svas3}. Specifically, after performing clustering on representative strain fields, each cluster will represent a ``sub-domain'' where sensors can then be placed at the medoid of the cluster.
Given a collection of sensors placed at the medoids for our homogeneous domain, we can then compute the reconstructed strain field. Namely, for each cluster, we replace the strain value of all the markers in the cluster with the strain value recorded at the medoid.

Looking forward, we anticipate that this approach might be relevant to soft robotics and related fields, where sensorization for proprioception is difficult due to continuum robots having a high number of degrees of freedom \citep{sensorsreview}. 
Currently, sensor positions are often selected based either on expert intuition \citep{wall2017method}, or locations determined with an optimization algorithm that is formulated based on expert intuition  \citep{bacher2016defsense}. For example, Tapia et al. obtained the optimal set of sensors with the following procedure: select a starting sensor setup; iteratively add sensors to the setup; reconstruct the deformation and external force with sensor data; compare the reconstructed deformation-force pairs to the FEM simulated deformation-force pairs, find the highest performing sensor via an optimization formulation that includes an analytical gradient; validate on a fabricated bar undergoing bending \citep{tapia2020makesense}. Alternatively, Spielberg et al. \citep{spielberg2021co} used a Particle Sparsifying Feature Extractor - a neural network trained to reduce dense sensor readings to a sparse representation of the sensor readings - alongside other common neural network architectures to learn an optimal sparse selection of sensors placements.
Though powerful, in depth optimization approaches may be either too challenging to implement for a given application, or ultimately unnecessary. 
And, though convenient, expert intuition may lead to ultimately inefficient sensor placements. 
Here, our goal is to explore unsupervised clustering as a sensor placement technique. Because this approach does not require either abundant training data or a high fidelity computational model of the physical system in question, it is a good baseline method to explore. In other words, we anticipate that this method will have acceptable performance, thus approaches that are more challenging to implement should be able to meaningfully outperform our method.
In the remainder of this Section, we will explore clustering for the purpose of informing sensor placement for strain field reconstruction. 

\begin{figure}[p]
\centering
\includegraphics[width=0.9\textwidth]{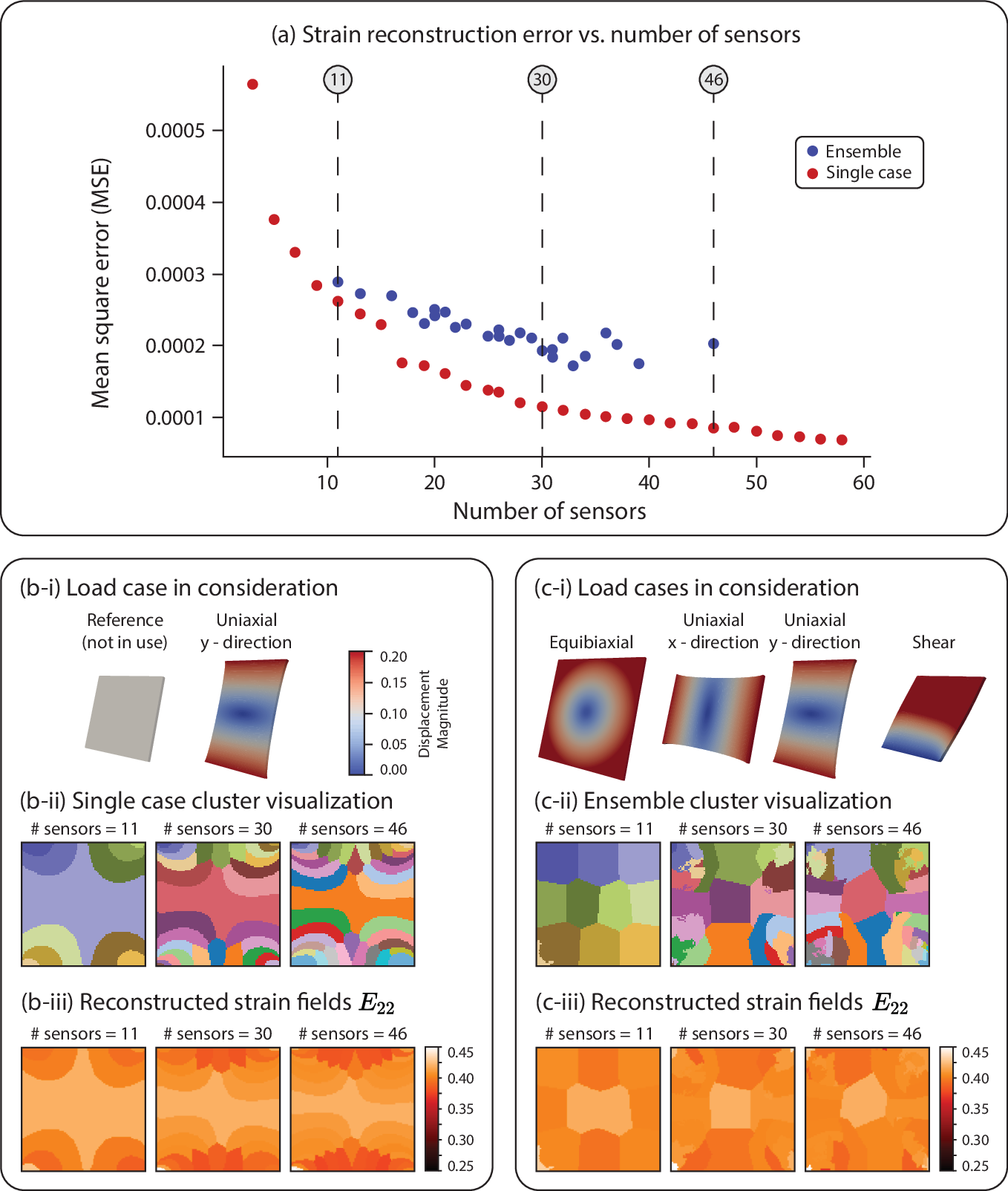}
\caption{Results showing the reconstructed strain fields $E_{22}$ for uniaxial extension in the y-direction: (a) the convergence plot showing the decrease in strain reconstruction error as the number of sensors increases; (b) illustration of clustering with a single load case, where (b-i) shows the load case in consideration, (b-ii) shows the cluster results visualization as the number of sensors increases for a single load case, and (b-iii) shows the reconstructed strain fields $E_{22}$ for uniaxial extension in the y-direction; (c) illustration of clustering using multiple loading cases, where (c-i) shows the four load cases that contribute to the ensemble, (c-ii) shows ensemble cluster results visualization as the number of sensors increases for multiple loading cases, and (c-iii) shows the reconstructed strain fields $E_{22}$ for uniaxial extension in the y-direction using multiple loading cases.}
\label{fig:MSEvnumsensors}
\end{figure}

To set up this investigation, we consider a homogeneous domain undergoing multiple modes of deformation (see Fig. \ref{fig:MSEvnumsensors}c). Given a strain field, we cluster the domain into multiple sub-domains, and consider a sensor placed at the medoid of each sub-domain. Then, we create the reconstructed strain field (details in Section \ref{sec:prob_def}) by borrowing a technique from the image compression literature \citep{kanungo2002efficient}. To evaluate the performance of our sensor placement suggestion, we compare the reconstructed strain field to the original strain field using mean squared error (MSE), where we take the average squared difference between the reconstructed strain and the original strain for all markers. 
For our clustering method, we choose \emph{k}-means and spectral clustering based on their top performance in Section \ref{sec:result1}. Here, we define our kinematic features via the Green-Lagrange strain tensor. Through these selections (i.e., \emph{k}-means, spectral clustering, and the Green-Lagrange strain), we reconstruct the strain fields, and compare the reconstructed strain fields for $2$ cases: the single loading case and the ensemble case. In the ensemble case, we cluster the domain by considering $4$ load cases, detailed below.

To generate data with a known ground truth for our investigation, we follow the process in Fig. \ref{fig:overview_dataset}d, and select a \emph{homogeneous} neo-Hookean square domain with side length $1$. For the single loading case (Fig. \ref{fig:MSEvnumsensors}b), we obtain the displacement field for the uniaxial extension boundary condition, and calculate the strain values for the grid markers following the process in Fig. \ref{fig:features}. Similarly, for our ensemble clustering process, we obtain the strain data from $4$ different boundary conditions (i.e., equibiaxial extension, uniaxial extension in the x-direction, uniaxial extension in the y-direction, shear, see Fig. \ref{fig:MSEvnumsensors}c). 
For the single load case example, we perform \emph{k}-means clustering and connected components labeling on our domain (process in Fig. \ref{fig:pipeline}a, results in Fig. \ref{fig:MSEvnumsensors}b). The single loading case is similar to many existing sensorization techniques, in that these techniques only consider a specific mode of deformation per sensors placements scheme \citep{svas3}. 
Then, to generalize the sensors placements scheme to account for multiple possible deformations, we perform ensemble clustering on the strain data from $4$ boundary conditions (process in Fig. \ref{fig:pipeline}, results in Fig. \ref{fig:MSEvnumsensors}c). 

In this example application, we anticipate that as the number of sensors increases, the resulting MSE will decrease. Intuitively, this is because as the number of clusters increases, the clusters will become smaller, and the medoid of each cluster will have a closer strain value to all other markers in the cluster, and the reconstructed strain field will become more similar to the original strain field. In Figure \ref{fig:MSEvnumsensors}a, we obtain the MSE for the single loading case, which shows a clearly decreasing trend as we expected. For the ensemble reconstructed strain, we observe a similar relationship between the number of sensors and the MSE. However, the single loading case converged at a lower MSE compared to the ensemble. Since the error is only evaluated on one load case (i.e., uniaxial extension in the y-direction), the result favors the single load case example. Despite the higher MSE for the ensemble, we believe that the ensemble clustering pipeline will provide more generalizable sensors placements suggestion that will be more useful to other boundary conditions. 

Overall, we provide a simple method to recommend sensors placements using only the Green-Lagrange strain, requiring little to no domain expertise, and accounting for multiple deformation modes. We hope that our method serves as a baseline for future sensors placements techniques to compare against.

\section{Conclusion} \label{sec:conclusion}

In this paper, we explore unsupervised learning as a tool to cluster unlabeled kinematic data from soft materials undergoing large deformation. Specifically, we evaluate unsupervised learning as a tool to both identify self-similar sub-domains within a heterogeneous domain, and identify self-similar regions within a heterogeneous strain field from a homogeneous material domain. 
To perform these studies, we extended our Mechanical MNIST dataset collection to include the Mechanical MNIST - Unsupervised Learning dataset, which simulates the behaviors of soft tissues and provides the \emph{in silico} data necessary to assess the performance of unsupervised methods. While our previous datasets contain a small amount of information for a large number of samples, this dataset contains a large amount of information for each individual sample. And, our unsupervised learning dataset also provides a ground truth, which is typically missing in the context of soft tissues, so we are able to quantitatively compare multiple different unsupervised learning techniques.

With this new dataset, we test 4 different methods from the unsupervised learning and anomaly detection literatures - \emph{k}-means clustering, spectral clustering, iForest clustering, and One-class SVM - on our circle inclusion sample undergoing different types of controlled boundary conditions. Here, we found that both \emph{k}-means and spectral clustering perform best (i.e., lead to the highest ARI scores). 
Next, we combine ensemble clustering and positional segmentation techniques from image analysis to create a more robust clustering pipeline.
By testing the clustering pipeline on 6 different heterogeneous patterns, we found that our new pipeline outperforms standard \emph{k}-means for heterogeneous samples undergoing controlled boundary conditions.
While our method works well for controlled boundary conditions, \emph{in vivo} experiments often have complex and asymmetrical boundary conditions. To assess the performance of our pipeline under more complex settings, we implement and test our clustering pipeline on samples undergoing random boundary conditions. Here, we found that for samples where the sub-domains have different properties for the ground substance, our clustering pipeline successfully identify the self-similar sub-domains. However, our clustering pipeline fails when the heterogeneous pattern is too complicated (i.e., Cahn-Hilliard), or when the ground substance remains the same across sub-domains.
Aside from soft tissue mechanics applications, we are also able to use our clustering pipeline to identify self-similar regions within a domain, and reconstruct the strain field. Here, we compare the performance of the clustering result using only a single boundary condition against the clustering result using an ensemble of multiple boundary conditions. We found that while the single boundary condition result provides a better reconstructed strain field when evaluated only on 1 strain field, the ensemble clustering result tend to be more generalizable.

In the future, we anticipate that our Mechanical MNIST - Unsupervised Learning dataset and our clustering pipeline will motivate multiple new research directions. To enable other researchers to build on our work, we have made our dataset and clustering pipeline available with a detailed tutorials (see Section \ref{sec:add_info} for access information). For soft tissue mechanics applications, we anticipate that our clustering pipeline will be useful in identifying mechanically self-similar sub-domains for scenarios where a region of soft tissue may experience multiple modes of unknown deformation. For applications in continuum soft robotics and similar applications, we have provided a baseline method to reconstruct the strain field with no additional information aside from the full-field strain of the domain. We anticipate that more sophisticated reconstruction methods will outperform our clustering pipeline given more information. Looking forward, we hope that the findings in this work will make unsupervised learning and heterogeneous materials simulations more accessible to both researchers and the general public and serve as a baseline for future methodological approaches.


\section{Additional Information} \label{sec:add_info}

The Mechanical MNIST - Unsupervised Learning dataset is available through the OpenBU Institutional Repository \url{https://open.bu.edu/handle/2144/46508} under a CC BY-SA 4.0 license \citep{https://open.bu.edu/handle/2144/46508}. With this dataset, we provide an abstract that describes the general purpose of the dataset, a supplementary document that details dataset structure, code to reproduce the dataset, and a tutorial with comprehensive instructions on utilizing the dataset. The code to reproduce both dataset generation via FEniCS and the clustering pipeline detailed in this paper are available on GitHub \url{https://github.com/quan4444/cluster_project} under a MIT License. 

\section{Acknowledgements} 
\label{sec:ack}

We would like to thank the staff of the Boston University Research Computing Services and the OpenBU Institutional Repository (in particular Eleni Castro and Yumi Ohira) for their invaluable assistance with generating and disseminating the ``Mechanical MNIST -- Unsupervised Learning Dataset.'' This work was made possible through start up funds from the Boston University Department of Mechanical Engineering, the David R. Dalton Career Development Professorship, the Hariri Institute Junior Faculty Fellowship, the Office of Naval Research Award N00014-22-1-2066, and the Office of Naval Research Award N00014-23-1-2450.

\begin{figure}[h]
\centering
\includegraphics[width=1\textwidth]{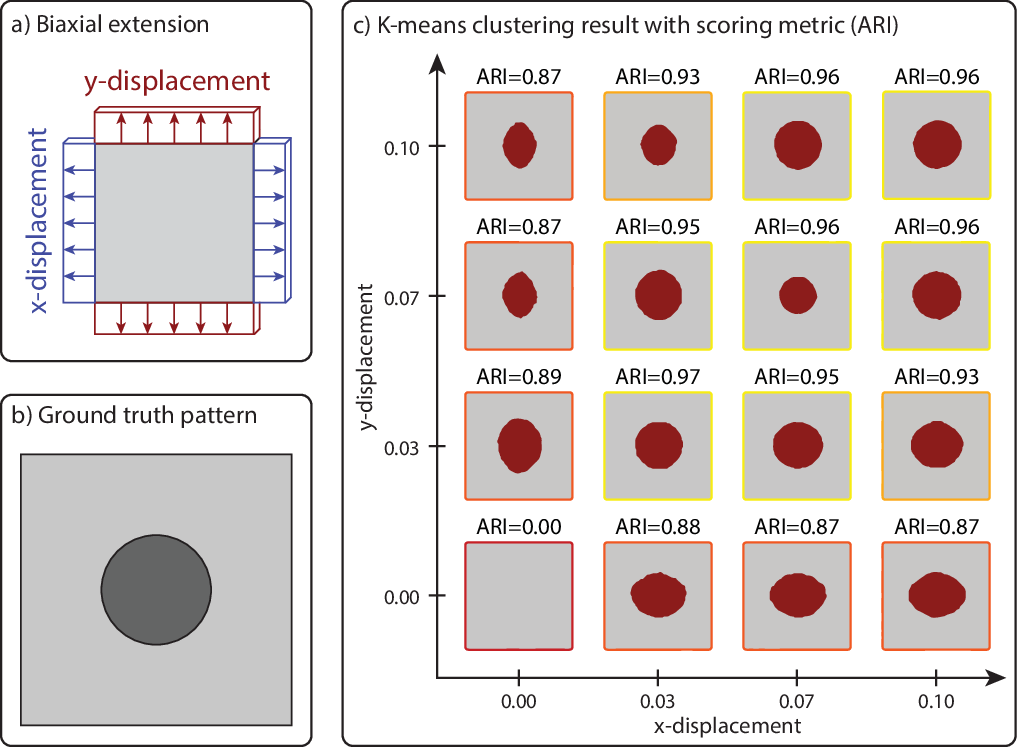}
\caption{\emph{K}-means clustering result for our circle inclusion pattern undergoing biaxial extension: (a) illustration of the biaxial extension load case with varying x and y displacements; (b) ground truth pattern used to evaluate the clustering results; (c) \emph{k}-means clustering result for different biaxial extension boundary conditions along with our scoring metric (ARI), where yellow boxes depict good clustering results, and red boxes depict poor clustering results.}
\label{fig:biaxial}
\end{figure}

\appendix 

\section{\emph{K}-means clustering result for circle inclusion with boundary conditions varying from equibiaxial to biaxial to uniaxial extension} \label{app:varying_bcs}

In Fig. \ref{fig:invariants}, \ref{fig:singlevsensemble}, and \ref{fig:random_bcs}, we show the results of applying our clustering pipeline to samples with equibiaxial, uniaxial in x, uniaxial in y, shear, and confined compression boundary conditions. In this Appendix, we provide an additional supplementary result where we vary the x and y displacements such that the boundary conditions fall between uniaxial and equibiaxial extension. In Fig. \ref{fig:biaxial}, we show the results of applying \emph{k}-means clustering to a circle inclusion with these varying boundary conditions. In brief, this result shows that \emph{k}-means clustering works well for a wide range of different biaxial extensions in the experimental setting. Overall, we observe that equibiaxial extension and near-equibiaxial extension boundary conditions provide the best clustering results (i.e., ARI $\geq 0.95$), while boundary conditions closer to uniaxial extension provide worse clustering results (i.e., ARI $< 0.90$). Despite the poorer performance in uniaxial extension cases, \emph{k}-means still identifies the circle inclusion quite well with the lowest $ARI=0.88$. Finally, as expected, \emph{k}-means fails to provide any reasonable results when the sample experiences no deformation.

\FloatBarrier
\newpage

\bibliography{references}  






\end{document}